\documentclass[journal]{IEEEtran}
\ifCLASSINFOpdf
\else
\fi
\newcommand{\mon}[1]{%
   \expandafter\monAUX\expandafter{\detokenize{#1}}%
}

\newcommand*\patchAmsMathEnvironmentForLineno[1]{%
\expandafter\let\csname old#1\expandafter\endcsname\csname #1\endcsname
\expandafter\let\csname oldend#1\expandafter\endcsname\csname end#1\endcsname
\renewenvironment{#1}%
{\linenomath\csname old#1\endcsname}%
{\csname oldend#1\endcsname\endlinenomath}}%
\newcommand*\patchBothAmsMathEnvironmentsForLineno[1]{%
\patchAmsMathEnvironmentForLineno{#1}%
\patchAmsMathEnvironmentForLineno{#1*}}%
\AtBeginDocument{%
\patchBothAmsMathEnvironmentsForLineno{equation}%
\patchBothAmsMathEnvironmentsForLineno{align}%
\patchBothAmsMathEnvironmentsForLineno{flalign}%
\patchBothAmsMathEnvironmentsForLineno{alignat}%
\patchBothAmsMathEnvironmentsForLineno{gather}%
\patchBothAmsMathEnvironmentsForLineno{multline}%
}

\usepackage{amsmath}
\usepackage{graphicx}
\usepackage{lineno}
\usepackage{array}
\usepackage{longtable}
\usepackage[numbers]{natbib}
\usepackage[usenames,dvipsnames]{color}
\usepackage{amssymb}
\usepackage{textcomp}
\usepackage{url}
\usepackage{float}
\usepackage{multirow}
\usepackage{microtype}

\begin{document}
%
\title{Weakly Supervised Semantic Segmentation in 3D Graph-Structured Point Clouds of Wild Scenes}%
%
%

\author{Haiyan Wang, Xuejian Rong, Liang Yang, Jinglun Feng,
        Jizhong Xiao, and~Yingli Tian*~\IEEEmembership{Fellow,~IEEE}
\IEEEcompsocitemizethanks{
\IEEEcompsocthanksitem Haiyan Wang, Xuejian Rong, Liang Yang, and Jinglun Feng are with the Department
of Electrical Engineering, The City College of New York, New York,
NY, 10031.\protect\\
E-mail: \{hwang3,xrong,lyang1,jfeng1\}@ccny.cuny.edu
\IEEEcompsocthanksitem Jizhong Xiao and Yingli Tian (*Corresponding author) are with the Department of Electrical Engineering, The City College, and the Department of Computer Science, the Graduate Center, the City University of New York, New York, NY, 10031.\protect\\
E-mail: \{jxiao,ytian\}@ccny.cuny.edu
}
\thanks{This material is based upon work supported by the National Science Foundation under award number IIS-1400802.}}

\maketitle

\begin{abstract}
The deficiency of 3D segmentation labels is one of the main obstacles to effective point cloud segmentation, especially for scenes in the wild with varieties of different objects. To alleviate this issue, we propose a novel deep graph convolutional network-based framework for large-scale semantic scene segmentation in point clouds with sole 2D supervision.
Different with numerous preceding multi-view supervised approaches focusing on single object point clouds, we argue that 2D supervision is capable of providing sufficient guidance information for training 3D semantic segmentation models of natural scene point clouds while not explicitly capturing their inherent structures, even with only single view per training sample. 
Specifically, a Graph-based Pyramid Feature Network (GPFN) is designed to implicitly infer both global and local features of point sets and an Observability Network (OBSNet) is introduced to further solve object occlusion problem caused by complicated spatial relations of objects in 3D scenes. During the projection process, perspective rendering and semantic fusion modules are proposed to provide refined 2D supervision signals for training along with a 2D-3D joint optimization strategy.
Extensive experimental results demonstrate the effectiveness of our 2D supervised framework, which achieves comparable results with the state-of-the-art approaches trained with full 3D labels, for semantic point cloud segmentation on the popular SUNCG synthetic dataset and S3DIS real-world dataset.
\end{abstract}

\begin{IEEEkeywords}
Deep Graph Convolutional Network, Point Cloud, 3D Semantic Segmentation, Weakly Supervised
\end{IEEEkeywords}

%
\IEEEpeerreviewmaketitle

\ifCLASSOPTIONcompsoc
\IEEEraisesectionheading{\section{Introduction}\label{sec:introduction}}
\else
\section{Introduction}
\label{sec:introduction}
\fi

\IEEEPARstart{T}{he} \label{intro}
The last decade has witnessed advances in 3D data capturing technologies which have become increasingly ubiquitous and paved the way for generating high accurate point cloud data including sensors such as laser scanners, time-of-flight sensors including \textit{Microsoft Kinect} or \textit{Intel RealSense} device, structural light sensors (e.g. \textit{iPhone X} and \textit{Structure Sensor}), outdoor \textit{LiDAR} sensors, etc. 3D information could significantly contribute to fine-grained scene understanding. For instance, depth information could drastically reduce segmentation ambiguities in 2D images, and surface normal in 3D data could provide important cues of scene geometry. However, 3D data are typically formed with point clouds (geometric point sets in Euclidean space), which are represented as a set of unordered 3D points with or without additional information such as the corresponding RGB images. The 3D points do not conform to the regular lattice grids as in 2D images. Directly converting point clouds to 3D regular volumetric grids might bring computation intractability due to  unnecessary sparsity and high-resolution volumes. The work in PointNet~\cite{Qi:2017:PointNetDL} and PointNet++~\cite{Qi:2017:PointNetDH} have pioneered the use of deep learning for 3D point cloud processing with handling the permutation invariance problem, including reconstruction and semantic segmentation tasks. However, these methods still heavily depend on 3D aligned point-wise labels as strong supervision signals for training, which are difficult and cumbersome to prepare and annotate.

\begin{figure}
\begin{center}
\includegraphics[width=0.5\textwidth]{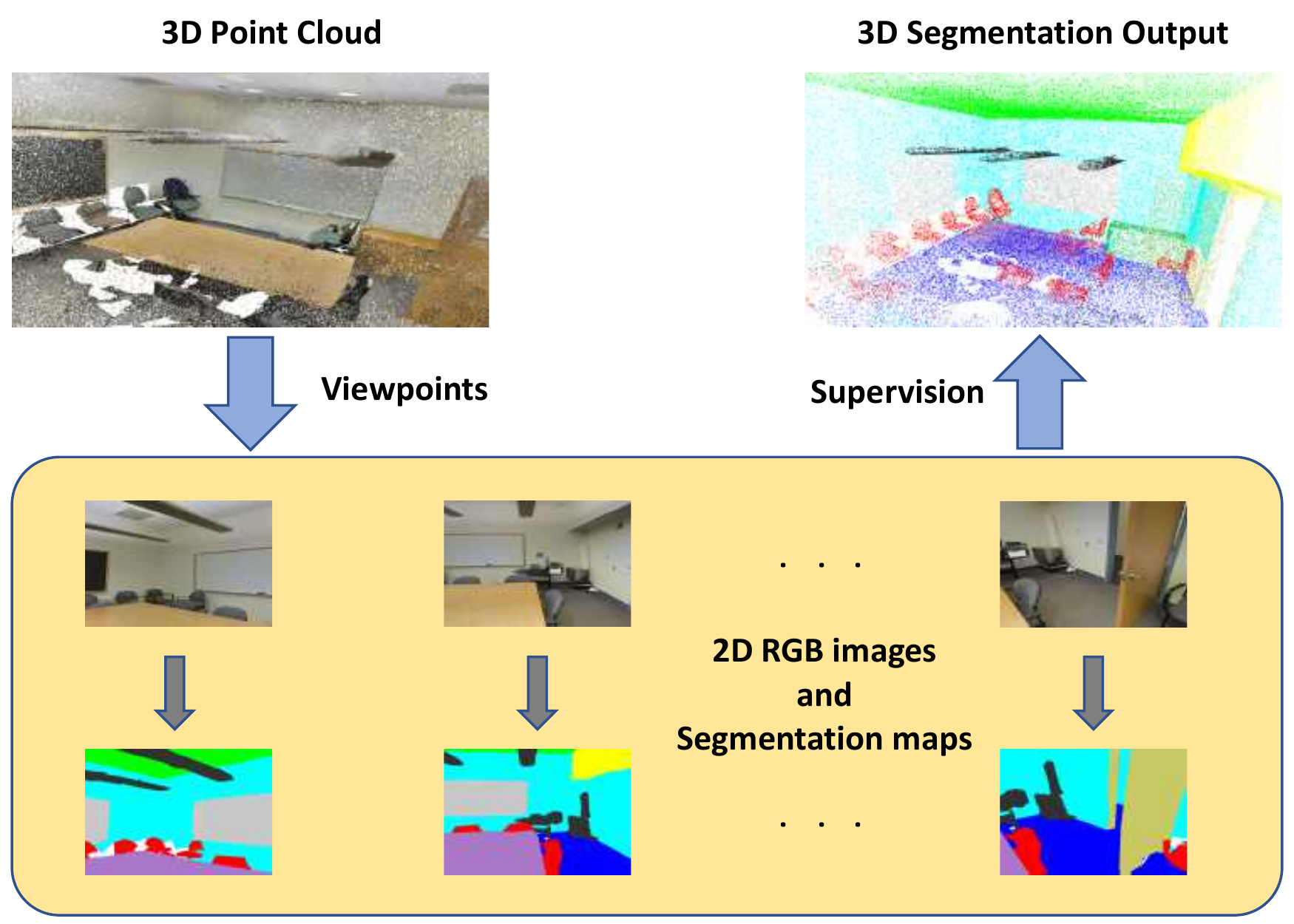}
\caption{Illustration of the proposed weakly 2D supervised semantic segmentation of 3D point cloud in the wild scenes. 
Without using point-wise 3D annotations, we leverage 2D segmentation maps of different viewpoints to supervise the 3D training process.}

\label{fig:motivation}
\end{center}
\end{figure}

\begin{table*}[ht!]
\centering
\caption{Definitions of the key terms used in the paper.}
\label{table:term}

\begin{tabular}{|c|l|}
\hline
\textbf{Term}                       & \multicolumn{1}{c|}{\textbf{Definition}}                                                                                                            \\ \hline
\textbf{Supervised Learning}        & \begin{tabular}[c]{@{}l@{}}learns mapping function between input and output pairs using fully labeled training examples.  \end{tabular}                                       \\ \hline  
\textbf{Weakly Supervised Learning} & \begin{tabular}[c]{@{}l@{}}learns mapping function between input and output pairs using  coarse or imprecise labels instead of \\fully labeled training examples.\end{tabular}    \\  \hline  
\textbf{Truncated Point Cloud}      & \begin{tabular}[c]{@{}l@{}}refers to the points inside a frustum under a specific viewpoint  in a 3D space. In our paper, it is obtained \\ by casting rays from the camera to the scene and extracting the points in a view (see Figure \ref{fig:truncated}) and used as \\ the input data to our framework.\end{tabular}   \\  \hline  
\textbf{3D Label \& 2D Label}       & \begin{tabular}[c]{@{}l@{}}3D label indicates the category label of each point for point cloud segmentation. 2D label refers to the \\ category label of each pixel in 2D segmentation maps.\end{tabular} \\ \hline
\end{tabular}
\end{table*}

Unlike existing methods which typically require expensive point-wise 3D annotations, as shown in the Figure \ref{fig:motivation}, this paper tackles the task of semantic point cloud segmentation for natural scenes by only utilizing popular 2D supervision signals such as 2D segmentation maps to supervise the 3D training process. We argue that 2D supervision is capable of providing sufficient guidance information to train 3D semantic scene segmentation models from point clouds while not explicitly capturing inherent structures of 3D point clouds. By rendering 2D  pixels from the point cloud, supervised by 2D segmentation maps, our proposed framework is able to learn semantic information for each point.  Compared to 3D data, 2D data are often much easier to obtain, thus save huge efforts to collect the ground truth label for each point in 3D supervision manner. Different with some recent 2D multi-view supervision-based single object 3D reconstruction approaches~\cite{Lin:2018:LearningEP, Liao20183DSR, Insafutdinov2018UnsupervisedLO} (enforcing cycle-consistency or not) which solely focus on single objects and require 2D data in multiple viewpoints, our approach works on the natural scene segmentation of point clouds for multiple objects with  only single view per truncated point cloud. 

Occluded objects may not be correctly labeled in generating 2D segmentation maps from a given viewpoint. Due to the sparseness of point cloud and unknown spatial relation and topology of surfaces in a scene, it is challenging to determine whether 3D points belong to occluded or visible objects by just using depth distances under specific camera viewpoints. As a result, if 3D point cloud is directly projected into 2D image planes, occluded points might also appear on images which results in a misguidance for the entire scene segmentation. Therefore identifying the spatial geometry relation of objects and removing such points from the projected 2D images are crucial to the design of joint optimization strategy.
In order to tackle the occlusion issue, we introduce an \emph{OBSNet}(Observability Network) to provide guidance for accurate projection of segmentation maps by removing the occluded points. Given a point cloud that contains  \texttt{RGB} and \texttt{Depth} information as input, the OBSNet directly outputs the visibility mask for each point. Furthermore, multiple points might collide if they are projected to same location in 2D images. Instead of simply using the depth attribute of points as a filtering mechanism, we propose a novel reprojection regime named \emph{perspective rendering} to perform semantic fusion for different points which significantly alleviates the point collision problem.



\begin{figure*}
\begin{center}
\includegraphics[width=1\textwidth]{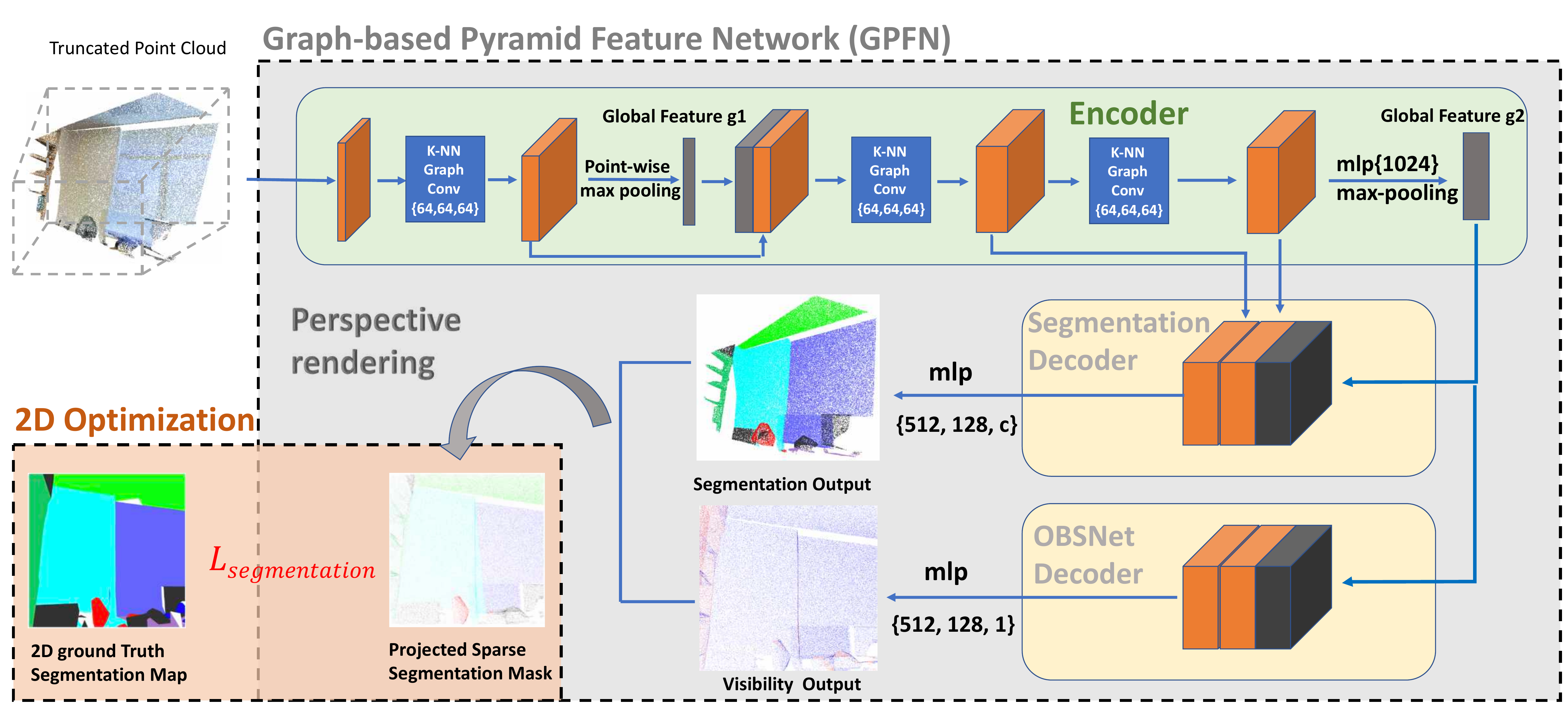}
\caption{The pipeline of the proposed deep graph convolutional framework for 2D-supervised 3D semantic point cloud segmentation. The GPFN network contains one encoder network and two decoder networks that share same encoder network. The first decoder is segmentation decoder to predict the segmentation point cloud. Another is OBSNet decoder to output the visibility of the point cloud. At last, the perspective rendering is designed to obtain the projected 2D mask which further jointly optimize the whole structure.}

\label{fig:new_pipeline}
\end{center}
\end{figure*}

The unified architecture illustrated in Figure~\ref{fig:new_pipeline} comprises a Graph-based Pyramid Feature Network (GPFN), a 2D perspective rendering module, and a 2D-3D joint optimizer. Specifically, the graph convolutional feature pyramid encoder works to hierarchically infer the semantic information of a scene in both local and global levels. The 2D perspective rendering works along with the predicted segmentation maps and the visibility masks to generate effective refined 2D maps for loss computation. The 2D-3D joint optimizer supports a complete end-to-end training. To make this paper easy to understand, we   define the key terms in Table \ref{table:term}.

\begin{figure}
\begin{center}
\includegraphics[width=0.5\textwidth]{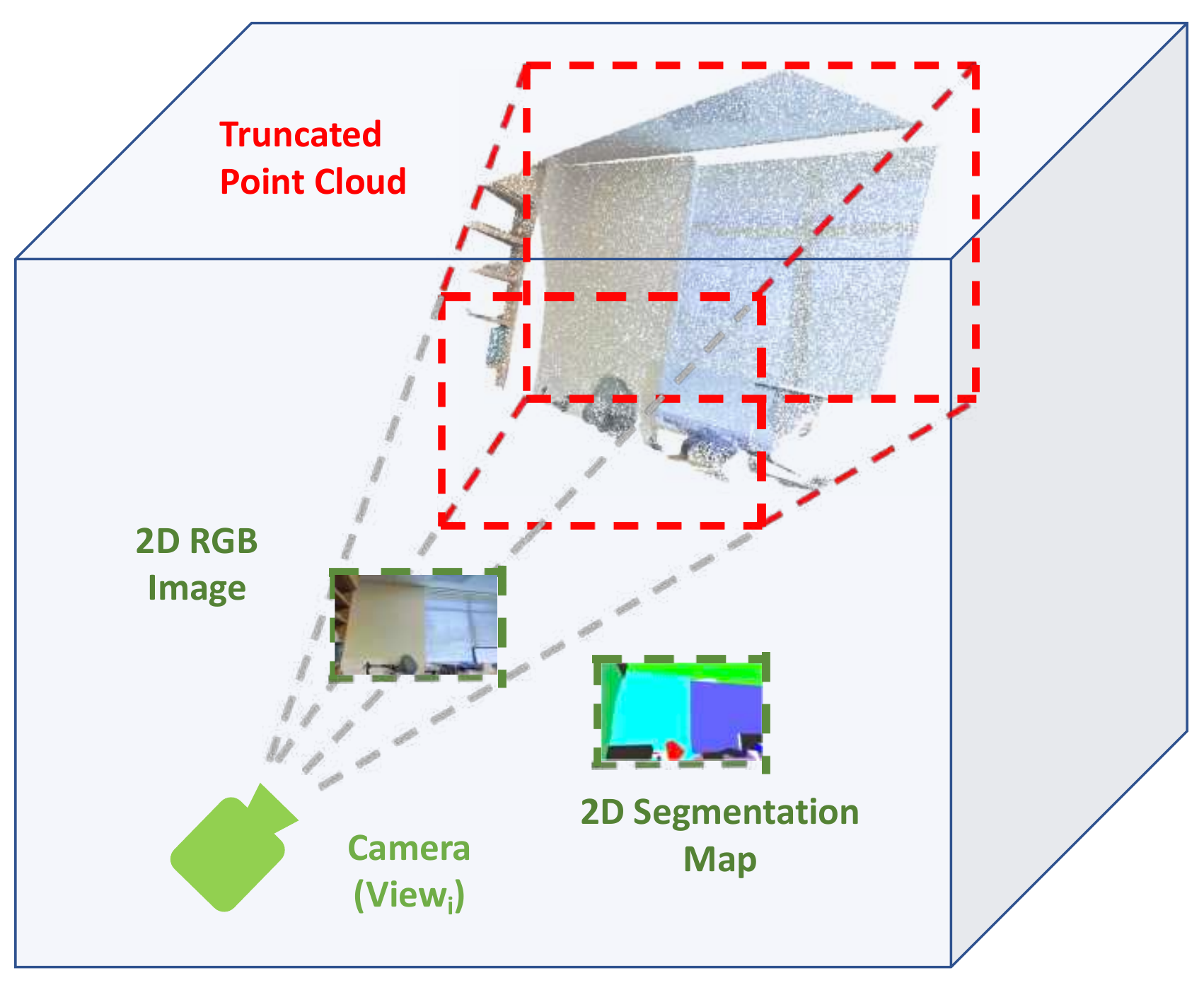}
\caption{Illustration of a truncated point cloud. The gray dashed lines refer to the rays casting from the camera and the area contained in the red dashed lines are the truncated point cloud under a viewpoint $v_i$. Note that there is one 2D RGB image corresponding to the truncated point cloud under the same viewpoint.} 

\label{fig:truncated}
\end{center}
\end{figure}

In an extension to our preliminary work~\cite{HaiyanBMVC}, instead of using the distance filter to solve the object occlusion problem, we introduce an OBSNet to our framework which learns to predict the visibility mask in an end-to-end manner. In addition, we explore the transfer learning from synthetic data to real-world data for 3D point cloud segmentation task. The main contributions are summarized as follows:
\begin{itemize}
  \item A joint 2D-3D deep architecture is designed to compute hierarchical and spatially-aware features of point clouds by integrating graph-based convolution and pyramid structure for encoding,  which further compensates weak 2D supervision information. 
  \item A novel re-projection method, named \textit{perspective rendering}, is proposed to enforce  2D and 3D mapping correspondence. Our approach significantly alleviates the needs for 3D point-wise annotations in training, while only 2D segmentation maps are used to calculate loss with the re-projection. 

  \item An observability network is introduced to predict if a point is visible or occluded and to generate a visibility mask without using any additional geometry information.  Combined with the segmentation map and the \textit{perspective rendering}, we can further take advantages of the 2D information to supervise the whole training process. 
  \item To the best of our knowledge, this is the first work to apply 2D supervision for 3D semantic point cloud segmentation of wild scenes without using any 3D point-wise annotations. Extensive experiments are conducted and the proposed method achieves comparable performance with the state-of-the-art 3D supervised methods on the popular SUNCG~\cite{Song2017SemanticSC} and S3DIS~\cite{Armeni20163DSP} benchmarks.
\end{itemize}

The rest of this article is organized as follows: 
Section 2 introduces related work in deep learning for 3D point cloud processing,
3D semantic segmentation, and 2D supervised methods for 3D tasks.
Section 3 describes the details of our framework for graph-based weakly supervised point cloud semantic segmentation.
Section 4 presents the datasets and experiments to evaluate the proposed weakly segmentation model. Finally, Section 5 summarizes the proposed work and points the future directions.

\section{Related Work}
\label{sec:1}
\subsection{Deep Learning for 3D Point Cloud Processing.}
In deep learning era, early attempts at using deep learning for large 3D point cloud data processing usually replicated successful convolutional architecture by converting point sets to regular grid-like voxels~\cite{Brock2016GenerativeAD,Dai:2017:ScanNetR3,Maturana2015VoxNetA3,Chou2019AVA,Krivokuca2019AVA}, which extended 2D CNN to 3D CNN and integrated the volumetric occupancy representation. The main problem of voxel-based methods is the huge number of parameters in the network and the increased spatial resolution.
Other methods based on k-d-tree~\cite{Bithell2007EscapeFT} and Octree~\cite{Riegler2016OctNetLD,Garcia2019GeometryCF} were proposed to deal with the point cloud data, which hierarchically partitioned the 3D Euclidean space and indexing. However, it takes expensive computation cost to build the k-d-tree and can be hard to fit the dynamic situation compared to Octree. As for Octree, even if it is much more efficiency, but object or scene can only be approximated, and not fully represented.

End-to-end deep auto-encoder networks were also employed to directly handle the point clouds. Achlioptas et al. conducted the unsupervised point cloud learning by using PointNet~\cite{Qi:2017:PointNetDL} similar encoder structure and three simple fully-connected layers as the decoder network ~\cite{Achlioptas2017RepresentationLA}. Although the design is simple and straight-forward, the generation model could already reconstruct the unseen object point cloud. FoldingNet~\cite{Yang2017FoldingNetPC} improved the auto-encoder design by integrating the graph-based encoder and a folding-based decoder network, which is more powerful and interpretable to reconstruct the dense and complete single object.

Recently more emerging approaches were proposed to directly feed point clouds to networks with fulfilling permutation invariance including PointNet~\cite{Qi:2017:PointNetDL}, PointNet++~\cite{Qi:2017:PointNetDH} and Frustum PointNets~\cite{qi2017frustum}.  Meanwhile, graph convolution methods demonstrated their effectiveness to solve the point cloud problem. RGCNN and DGCNN \cite{Wang:2018:DynamicGC,Te:2018:RGCNNRG}  were proposed to construct the graph of points first and then utilize the graph convolution to extract features. Due to the embedded topology and the geometry information in the graph structure, these networks demonstrated the potential higher capability to process point cloud data and achieved considerable successful performance on 3D point cloud-based tasks such as classification~\cite{Su2015MultiviewCN, Shi2015DeepPanoDP, Qi2016VolumetricAM, Guerry2017SnapNetRC3}, detection \cite{Ding2019PointCS}, segmentation~\cite{Ye20183DRN}, reconstruction~\cite{Mandikal:2018:3DLMNetLE}, completion~\cite{Yuan2018PCNPC,Hu2019LocalFI}, and etc. 
This paper focuses on the task of 3D point cloud semantic segmentation for natural scenes.

\subsection{3D Semantic Segmentation.}
Before PointNet was proposed, early deep learning-based methods have already become popular in solving 3D semantic segmentation using voxel-based methods~\cite{Tchapmi2017SEGCloudSS,Meng2018VVNetVV}. Voxelized data help raw point cloud becomes ordered and structured, and can be further processed by the standard 3D convolution. 
SegCloud~\cite{Tchapmi2017SEGCloudSS} is an end-to-end 3D point cloud segmentation framework that predicts coarse voxels first using the 3DCNN with trilinear interpolation(TI), then fully connected Conditional Random Fields (FC-CRF) were employed to refine the semantic information on the points and accomplish the 3D point cloud segmentation task. Other methods such as ~\cite{Song2017SemanticSC} and ~\cite{Dai2018ScanCompleteLS} tackled the semantic scene completion from the 3D volume perspective, as well as explored the relationship between scene completion and semantic scene parsing. Song et al. are the first to perform the semantic scene completion using a single depth image as input \cite{Song2017SemanticSC}. They focused on context learning using the dilated-based 3D context module and thus well predicted the occupancy and semantic label for each voxel grid. However, the limitations of the volume-based 3D methods are that they sacrificed the representation accuracy and hard to keep high-frequency textures with limited spatial resources.

Recently, some methods were proposed  to handle 3D semantic segmentation from the perspective of points and take the point cloud data as input which are permutation invariant, and output the class label for each point \cite{Qi:2017:PointNetDL, Qi:2017:PointNetDH}. SPG was proposed as a graph-based method to handle large scale point clouds or super points ~\cite{Landrieu:2018:LargeScalePC}. They partitioned a 3D scan scene into super-points which are parts with simple shapes according to their geometry constrains. In conjunction with the encoded contextual relationship between points, they further increased the prediction accuracy of the semantic labels.  Frameworks proposed in papers \cite{Engelmann:2017:ExploringSC, Engelmann:2018:KnowWY} aimed to enlarge the receptive field of the 3D scene and explored both the input-level and output-level context information for semantic segmentation. Also, a multi-scale architecture was applied to boost performance. Wang et al. proposed a method to find the promotion between instance segmentation and semantic segmentation \cite{Wang:2019:AssociativelySI}. The authors proved that the two tasks can be linked together and improved each other. Different than the existing methods, our approach focuses on effectively utilizing easily accessible 2D training data for 3D large-scale scenes.

\subsection{2D Supervision for 3D Tasks.}
While 3D supervised semantic segmentation has made great progress, many researchers started to explore using 2D labels to train networks for 3D tasks to reduce the heavy workload of labeling 3D annotations (point clouds, voxels, meshes, etc.), albeit most of which are designed for single objects. The work proposed in ~\cite{Lin:2018:LearningEP} attempted to generate point clouds for object reconstruction and applied a 2D projection mask and depth mask for joint optimization. The authors introduced a pseudo-rendering in the 2D image plane, which solves the collision within a single object during projection. However, the simple up-sampling followed with a max-pooling strategy only works well with a single object. When dealing with a more complex scene that contains multiple objects, the pseudo-rendering cannot guarantee to assign correct labels for different objects when they have collision.

NavaneetK et al.~\cite{NavaneetK:2019:CAPNetCA} proposed CAPNET for 3D point cloud reconstruction. The authors introduced a continuous approximation projection module and proposed a  differentiable point cloud rendering to generate a smooth and accurate point cloud projection. Through the supervision of 2D projection, their method achieved better reconstruction results compared to pseudo-rendering
~\cite{Lin:2018:LearningEP} 
and showed generalizability on the real data.

Chen et al. \cite{Chen2019PointBasedMS} proposed a network to predict depth images from point cloud data in a coarse-to-fine manner.  On the one hand, they directly predicted the depth image through the encoder-decoder network. On the other hand, they reproject the depth image to the 3D point cloud and calculated the 3D flow for each point. Combined with the 3D geometry prior knowledge and the 2D texture information, the network could iteratively refine the depth image with the ground truth and aggregate the multi-view image features.

Pittaluga et al. \cite{Pittaluga2019RevealingSB}  tackled the privacy attack task and reconstructed the RGB image from the sparse point cloud. The network took point cloud as input for a model contains three cascade Unets, and output the refined RGB image. Combined with RGB, depth, and SIFT descriptors, the first Unet estimated the visibility of each cloud point. Then the following two Unets, CoarseNet and RefineNet, are used to generate the coarse-to-fine RGB images. Novel views can also be generated by taking the virtual tours for the total scene. 

Follow the track of our preliminary work~\cite{HaiyanBMVC}, there are papers starting to explore the methods of applying the 2D supervision signal on the 3D scene point cloud segmentation task. Wang et al. \cite{Wang2019LDLS3O} proposes a method which conducts the 2D RGB image segmentation first using Mask R-CNN \cite{He2017MaskR}, and then the 2D semantic labels are diffused to the 3D space. Through the geometry graph connection with points, they finally obtain the semantic labels for the lidar point cloud. However, this paper heavily relies on the 2D segmentation neural networks such as Mask R-CNN and they didn't take advantage the global features of the point cloud. 

This paper extends our previous work \cite{HaiyanBMVC} and  proposes an unprecedented method towards better 2D supervision for 3D point cloud semantic scene segmentation and demonstrate its effectiveness on SUNCG synthetic dataset and S3DIS real-world dataset.

\begin{figure*}
\begin{center}
\includegraphics[width=1\textwidth]{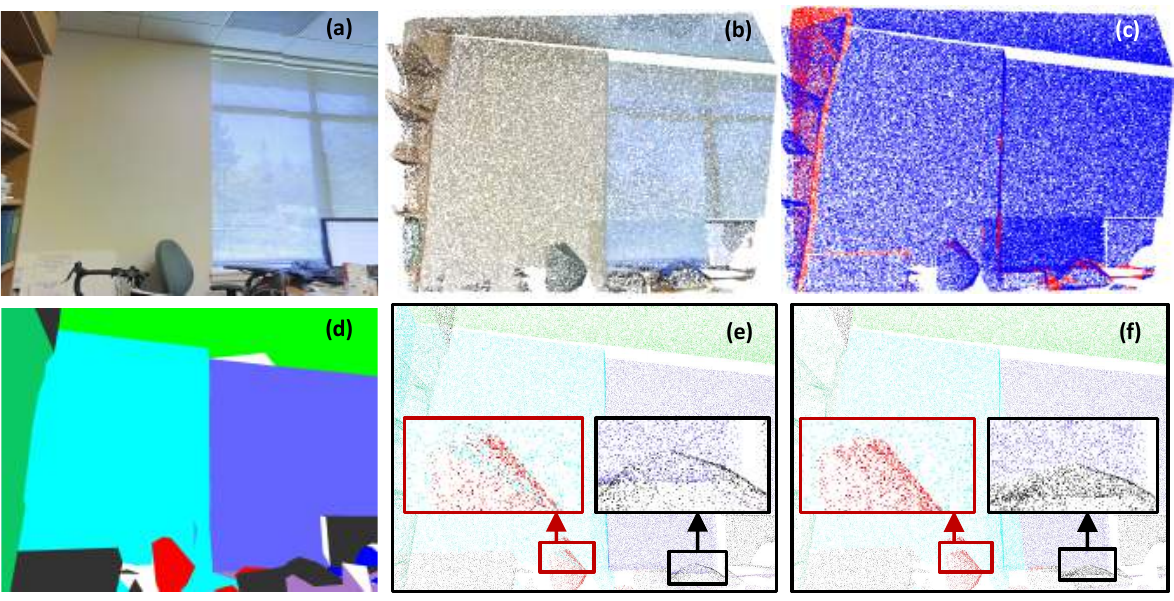}
\caption{Illustration of the effectiveness of visibility prediction by our proposed OBSNet.  (a) RGB image (just for visualization, and not used in training); (b) Truncated point cloud used as input to our network; (c) Visibility point cloud from the OBSNet;  (d) 2D ground truth segmentation map under same viewpoint; (e) the projected mask without OBSNet, and (f) the projected mask after applying the OBSNet. Two areas of point cloud semantic segmentation results with collision are zoomed in to visualize better details: in the red box, points of different objects (chair and wall) are projected in the same region in the 2D image plane before adding the OBSNet. After applying the visibility calculation, they are correctly separated. The collision problem is also resolved as shown in the corresponding black boxes between the clutter (in black) and window (in purple).}
\label{fig:pro_vs_gt}
\end{center}
\end{figure*}

\section{Methodology}
\label{sec:Methodology}

\subsection{Overview}

3D supervised deep models for semantic point cloud segmentation, such as PointNet~\cite{Qi:2017:PointNetDL}, PointNet++~\cite{Qi:2017:PointNetDH}, and DGCNN~\cite{Wang:2018:DynamicGC}, usually require 3D point-wise class labels in training and achieve satisfying results. To reduce the expensive labeling effort for each point in 3D point cloud data, here we propose a weakly 2D-supervised method by only using the 2D ground-truth segmentation map which is considerably easier to obtain to supervise the whole training process. Inspired by DGCNN~\cite{Wang:2018:DynamicGC}, we propose an effective encoder-decoder network to learn the representation of the point cloud. 


Figure~\ref{fig:new_pipeline} illustrates the proposed deep graph convolutional network-based framework for weakly supervised 3D semantic point cloud segmentation which comprises 2 main components: the Graph-based Pyramid Feature Network (GPFN) and the 2D optimization module. The \textbf{GPFN} takes the PointNet~\cite{Qi:2017:PointNetDL} similar structure as the baseline model which consists multiple MLP and max-pooling layers. Based on the baseline, the whole network contains a graph-based feature pyramid encoder and two decoder networks. A truncated point is obtained by casting rays from the camera through each pixel to the scene and extracting the points under a specific viewpoint (see details in Section~\ref{sec:gcnfp_encoder}.) The encoder takes a truncated point cloud from a given viewpoint as input. Then in order to solve the object occlusion problem, a novel framework with double-branch decoders  is designed. A segmentation decoder predicts the semantic segmentation labels while a visibility decoder  estimates the visibility mask for the scene point cloud under a specific viewpoint. The segmentation map and the visibility mask are further combined to handle the point collision problem and project a sparse 2D segmentation map. During \textbf{2D Optimization}, the projected sparse segmentation map is projected from the predicted segmentation point cloud by perspective rendering. Then the 2D ground truth segmentation map is finally applied to calculate the 2D sparse segmentation loss as the supervision signal in the training phase. To the best of our knowledge, this is the first work of applying weakly 2D supervision to the point cloud semantic scene segmentation task.

\subsection{Graph Convolutional Feature Pyramid Network} 
\label{sec:gcnfp_encoder}


By casting rays from the camera through each pixel to the scene, the points under specific viewpoints are extracted respectively to obtain the truncated point cloud for multiple viewpoints. An encoder-decoder network ($E_{p}$, $D_{p}$) is trained which takes the truncated input point cloud of $X_{p}\in \mathbb{R}^{N\times6}$ ($N$ is the number of the points and $6$ is the dimension of each point including $XYZ$ and $RGB$) from a given viewpoint $v = \{R, t\}$ and predicts the class label of the point cloud with size of $\widehat{X}_{p} \in \mathbb{R}^{N\times C}$ ($C$ is the number of classes.)

First, the truncated point cloud from a given viewpoint is fed into the encoder network $E_{p}$, which is comprised of a set of 1D convolution layers, edge graph convolution layers, and max pooling layers to map the input data to a latent representation space. Then the segmentation decoder network $D_{s}$ processes the feature vector through several fully-connected layers $(512, 128, C)$ and finally output the class prediction of each point.

In order to conjunct with the weak 2D labels, a graph-based feature pyramid encoder is designed to subside the effect of weak labels to the point cloud segmentation. Benefiting from the dynamic graph convolution model and pyramid structure design, the network could globally capture the semantic meaning of a scene in both low-level and high-level layers. Inspired by~\cite{Wang:2018:DynamicGC}, we introduce the K-NN dynamic graph edge convolution here. For each graph convolution layer, the K-NN graph is different and represented with $\mathcal{G}^{(l)} = {(\mathcal{V}^{(l)}, \mathcal{E}^{(l)})}$. ${|\mathcal{V}|}$ represents the nearest $k$ points to $x_{i}$, and $\mathcal{E}$ stands for edges between $(i, j_1)$, ... , $(i, j_k)$. Through the graph convolution $h_{\theta}$, the local neighborhood information is aggregated by capturing edge features between $k$ neighbors and center points: 
\begin{equation}
    \begin{aligned}
        h_{\theta}(x_i, x_j) = h_{\theta}(x_i, x_j-x_i).
    \end{aligned}
\end{equation}

As shown in Figure~\ref{fig:new_pipeline}, two pyramid global layers are added to the GPFN. The global features $g1, g2$ are concatenated with the previous point features in both low level and high level. This pyramid design and augmented point feature matrix are effective to improve the performance when using 2D supervision.


\subsection{Visibility Estimation}
In projection, the collision problem, namely the points on occluded objects and visible objects of various classes might be projected to the same location, which may cause intersections in the image plane. As shown in Figure~\ref{fig:pro_vs_gt}, there exist collisions such as between bookcase and wall, computer and window, chair and wall, etc. In order to explore the spatial location relation of points, we need to figure out which point should be considered as visible under the specific viewpoint. So the removal of such occluded points becomes crucial to our task. Otherwise, it would be difficult to accurately utilize the 2D supervision for point cloud segmentation. In our previous work~\cite{HaiyanBMVC}, we introduced the geometric-based method \emph{distance filter}, which requires additional effort such as calculating the boundaries of segmentation maps to solve the occlusion problem. In this paper, we propose an end-to-end network structure that contains the OBSNet Decoder $D_v$ to solve the occlusion problem based on the data-driven training.

In order to simplify the solution of finding the objects' spatial relationship and better solve the occlusion problem, we propose an end-to-end regression-based model to determine the visibility for the point cloud. As shown in Figure~\ref{fig:new_pipeline}, the OBSNet $D_v$ shares the same encoder network with the segmentation decoder $D_s$, taking the truncated input point cloud of $X_{p}\in \mathbb{R}^{N\times6}$ as input and outputs the label of "visible" or "occluded" for each point. The OBSNet decoder is also combined with both low level and high-level features that aggregate the geometric prior and spatial information, trained in a supervised manner and predicts the single label output $X_{p}\in \mathbb{R}^{N\times1}$. Therefore, the ground-truth visibility labels are obtained through the distance filter (see more details in ~\cite{HaiyanBMVC}) during both training and testing.
As a result, a point will be eliminated if it is classified as an occluded point and will not contribute to the loss calculation in the optimization process. 

During training, the two decoders could mutually help and benefit from each other. Furthermore, the OBSNet decoder helps the network separate the spatial location of objects based on the distance to the camera. To some extent, this provides a rough segmentation for the 3D scene. Overall, the segmentation model is able to learn enough semantic features and context information as guidance for the visibility prediction.

\begin{figure}
\begin{center}
\includegraphics[width=0.45\textwidth]{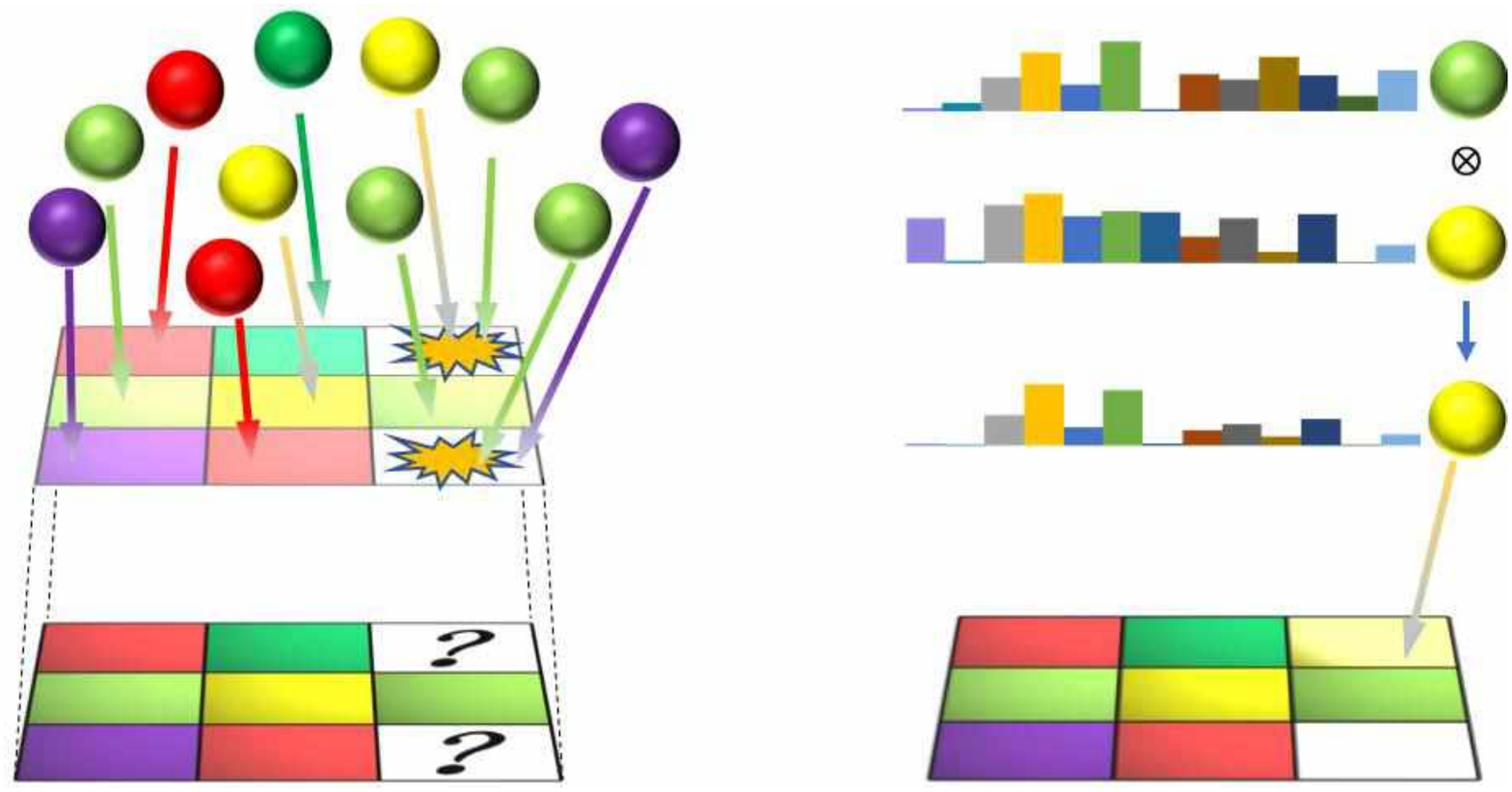}
\caption{Concept illustration of the proposed perspective rendering and semantic fusion. During the projection, multiple points of different object classes (shown in different colors) are projected to grids (with corresponding colors) in the image plane. Here, each grid indicates a pixel in the image. The left-side figure demonstrates the points collision problem which multiple points might be projected to the same grid. And we provide the solution in the right-side figure. Each point has a probability distribution of the predicted classes. For the grid which has multiple projected points, the perspective rendering is applied by calculating the dot product of the probabilities for all the points according to the classes, and after normalization, the class label of this grid can be finally determined.}
\label{fig:perspective}
\end{center}
\end{figure}

\subsection{Perspective Rendering}

For jointly optimizing the 2D and 3D networks and solving the point collision problem, we propose an innovative projection method named \emph{perspective rendering}. Point cloud in the world coordinate system is represented as $p_{w} = (x_{w}, y_{w}, z_{w})$. Camera pose and 3D transformation matrix for a given viewpoint are denoted as ($R_k$, $t_k$). The projected point in the camera coordinate system $\widehat{p}_{c} = (x_{c}, y_{c}, z_{c})$ can be derived through Eq.~\ref{equ: projection}.

\begin{equation}
\begin{aligned}
\label{equ: projection}
  \ \widehat{p}_{c} = (\widehat{x}_{c}, \widehat{y}_{c}, \widehat{z}_ {c}) &= (R_k p_w+t_k)\\ &= (R_k (x_{w}, y_{w}, z_{w})+t_k).
\end{aligned}
\end{equation}

However, as shown in Figure~\ref{fig:perspective}, different points might be projected to the same pixel position in the image plane. Through Eq.~\ref{equ: fusion}, the perspective rendering is applied for semantic fusion by predicting the probability distribution across all classes and fusing the probability of all the $N$ points which are projected to the same pixel position. At last, the probability distribution of this pixel is obtained through semantic fusion, the largest probability such as yellow shown in Figure~\ref{fig:perspective} is assigned as the final prediction label of this pixel.
\begin{equation}
\begin{aligned}
\label{equ: fusion}
  & p(C_i|x_{grid})  = \prod_{n=1}^{N}p(C_i|x_n), \quad \\
 &  p(C_i|x_{grid})_{norm}  = p(C_i|x_{grid})/\sum_{i=1}^{n_{classes}}\prod_{n=1}^{N}p(C_i|x_n),
  \\ & p(x_{grid})  = max\{p(C_1|x_{grid}), ..., p(x_{C_{n_{classes}}|grid})\}.
\end{aligned}
\end{equation}

\subsection{2D Optimization}

The ground-truth segmentation map $\hat{p_i}$ and the visibility mask $\hat{v_i}$ are used for enforcing the consistency among the prediction results. The loss function here contains the sparse point segmentation loss $L_{seg}$ and the visibility mask loss $L_{vis}$. 
The sparse loss is calculated for the projected segmentation result in training as the following equation: 
\begin{align}
\label{equ:loss}
  \ L_{seg} &= -\frac{1}{N}\sum_{i=1}^{N}
  {[
  p_i\log\hat{p_i} + (1-p_i)\log(1-\hat{p_i}))
  ]},\;
\end{align}
where $p_i$ is the predicted point cloud label projected to the 2D image plane. According to the 2D coordinates of the projected points, $\hat{p_i}$ is obtained by finding the labels of the corresponding points in the ground truth segmentation map.
\begin{align}
\label{equ:loss_vis}
  \ L_{vis} &= -\frac{1}{M}\sum_{i=1}^{M}
  {[
  U_i\log\hat{U_i} + (1-U_i)\log(1-\hat{U_i}))
  ]}.\;
\end{align}
$L_{vis}$ is using binary cross-entropy loss over $M$ non-zero valid predicting points which is similar to Eq.~\ref{equ:loss}. The total loss is calculated as $L = L_{seg} + \lambda L_{vis}$, where $\lambda$ is the weighting factor.

\section{Experiments}
\label{sec:Experiments}

\subsection{Datasets}

The proposed weakly 2D-supervised 3D point cloud semantic segmentation method is evaluated on two public and challenging 3D wild scene datasets, including 1)~\emph{SUNCG} ~\cite{Song2017SemanticSC}, a synthetic 3D large-scale indoor scene dataset, and 2) S3DIS (\emph{Stanford Large-Scale 3D Indoor Spaces}) dataset~\cite{Armeni20163DSP} derived from real environments.

\noindent \textbf{SUNCG Synthetic Dataset.}
SUNCG~\cite{Song2017SemanticSC} is a large-scale synthetic scene dataset that contains $45,622$ different indoor scenes with realistic rooms and furniture layouts that are manually created through the Planner5D platform. It  contains $404,058$ rooms and $5,697,217$ object instances. 

In this project, we create a total of $55,000$ 2D rendering sets. Each 2D rendering set comprises RGB images, depth images, and segmentation map with the corresponding camera viewpoints. The entire indoor scene point cloud can be obtained by back-projecting the depth images from every viewpoint inside a scene and fusing them together. Specifically, we only keep the rooms which have more than \textit{15} viewpoints and related rendered depth maps. There are total 40 object categories in the dataset including \textit{wall, floor, cabinet, bed, chair, sofa, table, door, window, bookshelf, picture, counter, blinds, desk, shelves, curtain, dresser, pillow, mirror, floor\_mat, clothes, ceiling, books, refrigerator, television, paper, towel, shower\_curtain, box, whiteboard, person, night\_stand, toilet, sink, lamp, bathtub, bag, otherstructure, otherfurniture, and otherprop}. The generated truncated point cloud data used in our training process and the 2D rendering sets will be released to the public along with the acceptance of this paper. 


\noindent \textbf{S3DIS Real-world Dataset.}
The Stanford Large-Scale 3D Indoor Spaces (S3DIS) dataset contains various larger-scale natural indoor environments and is significantly more challenging than other real 3D datasets such as ScanNet~\cite{Dai2018ScanCompleteLS} and SceneNN~\cite{Hua2016SceneNNAS} datasets. It consists of 3D scan point clouds
for $6$ indoor areas including a total of $272$ rooms. For each room, thousands of viewpoints are provided, including camera poses, 2D RGB images, 2D segmentation maps, and depth images under each specific viewpoint. For semantic segmentation, there are $13$ object categories including \textit{ceiling, floor, wall, beam, column, window, door, table, chair, bookcase, sofa, board,} and \textit{clutter}.  

\subsection{Implementation Details}

For both SUNCG and S3DIS datasets, each point is represented as a normalized flat vector (\texttt{XYZ, RGB}) with the dimension of $6$. These truncated point clouds are used as training data as well as calculating the loss with a 2D segmentation map under the same viewpoint. Following the settings as in ~\cite{Qi:2017:PointNetDL}, in which each point is represented as a 9D vector (\texttt{XYZ, RGB, UVW}), here \texttt{UVW} are the normalized spatial coordinates. In testing, the testing data are the points of the entire room similar to other 3D fully-supervised methods. For the SUNCG dataset, in total, $50,000$ viewpoints are selected and used to truncate point cloud as our training data, while with $5,000$ viewpoints as our testing data.  For S3DIS, The experimental results are reported by training on the $1/6$ viewpoints of training data (see details in Section \ref{sec:amount}) and testing on the 6-fold cross-validation over the 6 areas (\textit{area 1 - area 6}). 
Our proposed network is trained with $100$ epochs with batch size $48$, base learning rate $0.001$ and then is divided by $2$ for every $300k$ iterations. The Adam solver is adopted to optimize the network on a single GPU. A connected component algorithm is employed to calculate the boundary of each instance in the ground truth segmentation map.
The performance of semantic segmentation results is evaluated by the standard metrics: mean accuracy of total classes (\textit{mAcc}), mean per-class intersection-over-union (\textit{mIOU}), and overall accuracy (\textit{oAcc}).

\subsection{Experimental Results}

\begin{table*}
  \centering
  \setlength{\tabcolsep}{6pt}
 \caption{Quantitative results of our proposed 2D supervised method on SUNCG dataset. "w/" indicates "with" and "w/o" indicates "without". "DP" indicates Direct Projection, "PR" indicates Perspective Rendering and "$D_v$" indicates the OBSNet decoder.}
 \vspace{1pt}
  \begin{tabular}{c|c|c|c|c}
  \hline
    \multicolumn{1}{l|}{}               &
    \textbf{Method}                                & \textbf{mAcc(\%)} & \textbf{mIoU(\%)} & \textbf{oAcc(\%)} \\ \hline 
 
    \multirow{4}{*}{2D Supervision}    &  \textit{GPFN with DP (Ours)}    &      61.9         &     45.0           &    73.4    \\
                        &  \textit{GPFN with DP w/ $D_v$ (Ours)}    &      71.9         &     61.2        &   84.5    \\
                        &  \textit{GPFN with PR w/o $D_v$ (Ours)}    &     65.3          &     50.8         &   79.1    \\
                        & \multicolumn{1}{c|}{\textit{GPFN with PR w/ $D_v$ (Ours)}}       &   \multicolumn{1}{c|}{\textbf{87.3}}       &    \multicolumn{1}{c|}{\textbf{70.37}}     & \multicolumn{1}{c}{\textbf{91.8}}
    \end{tabular}
  \\[4pt]
 
  \label{table:comparison_suncg}
\end{table*}

\begin{table*}
  \centering
  \setlength{\tabcolsep}{6pt}
 \caption{Quantitative results without pretrained model of our proposed 2D supervised method on S3DIS dataset by using only 1/6 of viewpoints in each room for training. The performance of our 2D supervised method achieves comparable results with most of the 3D supervised state-of-the-art methods.}
 \vspace{1pt}
  \begin{tabular}{c|c|c|c|c}
  \hline
    \multicolumn{1}{l|}{}               &
    \textbf{Method}                                & \textbf{mAcc(\%)} & \textbf{mIoU(\%)} & \textbf{oAcc(\%)} \\ \hline 
    \multirow{5}{*}{3D Supervision}     & \textit{PointNet}~\cite{Qi:2017:PointNetDL}                              & 66.2          & 47.6          & 78.5          \\
                                        & \multicolumn{1}{c|}{\textit{Engelmann et al.}~\cite{Engelmann:2017:ExploringSC}}                      & \multicolumn{1}{c|}{66.4}          & \multicolumn{1}{c|}{49.7}          & \multicolumn{1}{c}{81.1}          \\
                                        & \textit{PointNet++}~\cite{Qi:2017:PointNetDH}                            & 67.1          & 54.5          & 81.0          \\
                                        & \multicolumn{1}{c|}{\textit{DGCNN~\cite{Wang:2018:DynamicGC}}}      & \multicolumn{1}{c|}{-}             & \multicolumn{1}{c|}{56.1}          & \multicolumn{1}{c}{84.1}           \\
                                        & \multicolumn{1}{c|}{\textit{Engelmann et al.}~\cite{Engelmann:2018:KnowWY}} & 67.8          & 58.3          & 84.0          \\
                                        & \multicolumn{1}{c|}{\textit{SPG}~\cite{Landrieu:2018:LargeScalePC}}                                   & \multicolumn{1}{c|}{73.0}          & \multicolumn{1}{c|}{62.1}          & \multicolumn{1}{c}{85.5}
                                        \\ \hline
    \multirow{4}{*}{2D Supervision}    &  \textit{GPFN with DP (Ours)}    &    39.2            &     30.4           &        53.7\\
                        &  \textit{GPFN with DP w/ $D_v$ (Ours)}    &     59.4          &    42.7         &   70.0    \\
                        &  \textit{GPFN with PR w/o $D_v$ (Ours)}    &     54.2          &      39.0         &    66.8    \\
                        & \multicolumn{1}{c|}{\textit{GPFN with PR w/ $D_v$ (Ours)}}                                  &   \multicolumn{1}{c|}{66.5}            &    \multicolumn{1}{c|}{50.8}           &       \multicolumn{1}{c}{79.1}
    \end{tabular}
  \\[4pt]
 
  \label{table:comparison_1}
\end{table*}

\subsubsection{Effectiveness of the Proposed Framework}

\noindent \textbf{2D Supervised-GPFN by Direct Projection without OBSNet.} 
Instead of using 3D ground truth labels as supervision, here only 2D segmentation maps are adopted for training. The predicted point cloud with labels is re-projected to the image by direct projection according to the camera model pose \textit{(R,t)}, while the loss is calculated based on the 2D segmentation maps. Note that point collision might occur while the occluded object is projected to the same area as the visible object. Not surprisingly, on both the synthetic and real-world datasets, as shown in the first row under "2D Supervision" in Table \ref{table:comparison_suncg} and Table \ref{table:comparison_1}, the performance ($61.9\%$ mAcc, $45.0\%$ mIoU, and $73.4\%$ oAcc) on SUNCG and ($39.2\%$ mAcc, $30.4\%$ mIoU, and $53.7\%$ oAcc) on S3DIS are extremely low. 

\noindent \textbf{2D Supervised-GPFN by Direct Projection with OBSNet.}

We conduct experiments by adding the OBSNet decoder $D_v$  but still using the direct projection mentioned before. Even the point collision problem still exists, the spatial relation between the visible and occluded objects are distinguished through the $D_v$. This is especially important in 3D scenes when there are multiple classes of objects. Thus on the SUNCG dataset, the performance is boosted up to ($71.9\%$ mAcc, $61.2\%$ mIoU, and $84.5\%$ oAcc) (the 2nd row in Table \ref{table:comparison_suncg}) and on S3DIS dataset, the performance is boosted up to ($59.4\%$ mAcc, $42.7\%$ mIoU, and $70.0\%$ oAcc) (the 2nd row in Table \ref{table:comparison_1}), which demonstrate the huge positive impact of the proposed OBSNet. 

\noindent \textbf{2D Supervised-GPFN by Perspective Rendering without OBSNet.} 
We further explore the effectiveness of the perspective rendering. During the network design, we only keep the segmentation decoder $D_s$ and perform the semantic fusion when projecting the point cloud to the 2D image plane. In this way, the points inside each single object might be well predicted via fusion. However, for the complex scenes and multiple objects, due to the occlusion issue, the improvement is limited with the performance of ($54.2\%$ mAcc, $39.0\%$ mIoU, and $66.8\%$ oAcc) on S3DIS dataset. For the SUNCG dataset, the environments in the dataset are complicated and the occlusion issue would more frequently occur. Semantic fusion cannot contribute much therefore the performance is only improved to ($65.3\%$ mAcc, $50.8\%$ mIoU, and $79.1\%$ oAcc) on SUNCG dataset.

\noindent \textbf{2D Supervised-GPFN by Perspective Rendering with OBSNet.} 
Our proposed \textit{Perspective Rendering} replaces the direct projection in this experiment. 
Combined with the OBSNet, the predicted point cloud is filtered via the visibility mask. Furthermore, the points that are projected to the same grid are selected by semantic fusion. For the synthetic dataset, since there is no other method conducting point cloud segmentation, we only compare the result among different architectures of our proposed GPFN. As shown in Table~\ref{table:comparison_suncg}, the result is largely improved to ($87.3\%$ mAcc, $70.37\%$ mIoU, and $91.8\%$ oAcc) due to the associate impacts between \textit{Perspective Rendering} and OBSNet. For the real-world dataset, as shown in Table~\ref{table:comparison_1}, the segmentation results ($66.5\%$ mAcc, $50.8\%$ mIoU, and $79.1\%$ oAcc) are significantly improved and even comparable with fully 3D-supervised results.

\begin{figure*} 
\begin{center}
\includegraphics[width=1\textwidth]{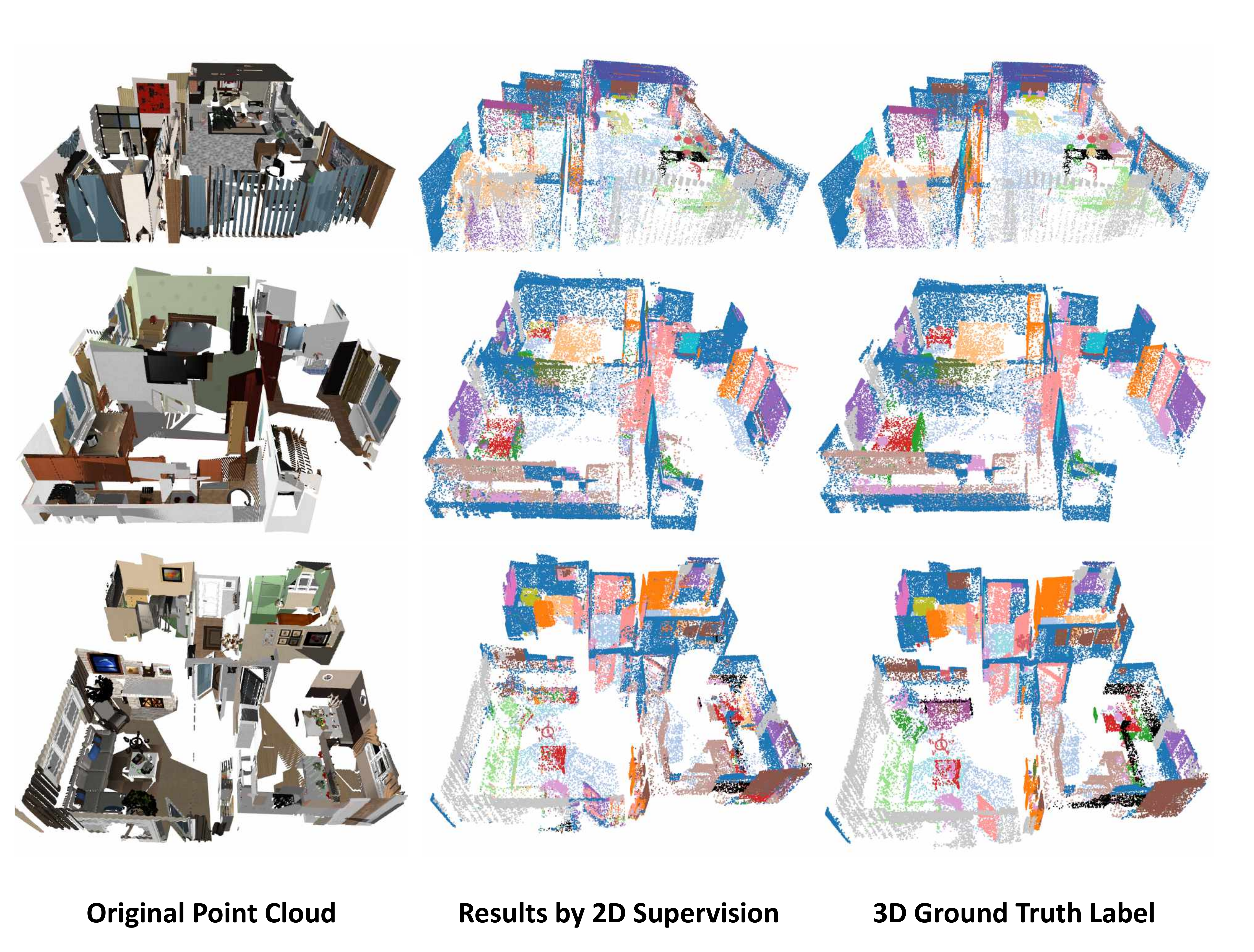}
\caption{Qualitative results produced by our proposed method (middle column) on SUNCG dataset.}
\label{fig:results_suncg}
\end{center}
\end{figure*}


\begin{figure*}
\begin{center}
\includegraphics[width=0.95\textwidth]{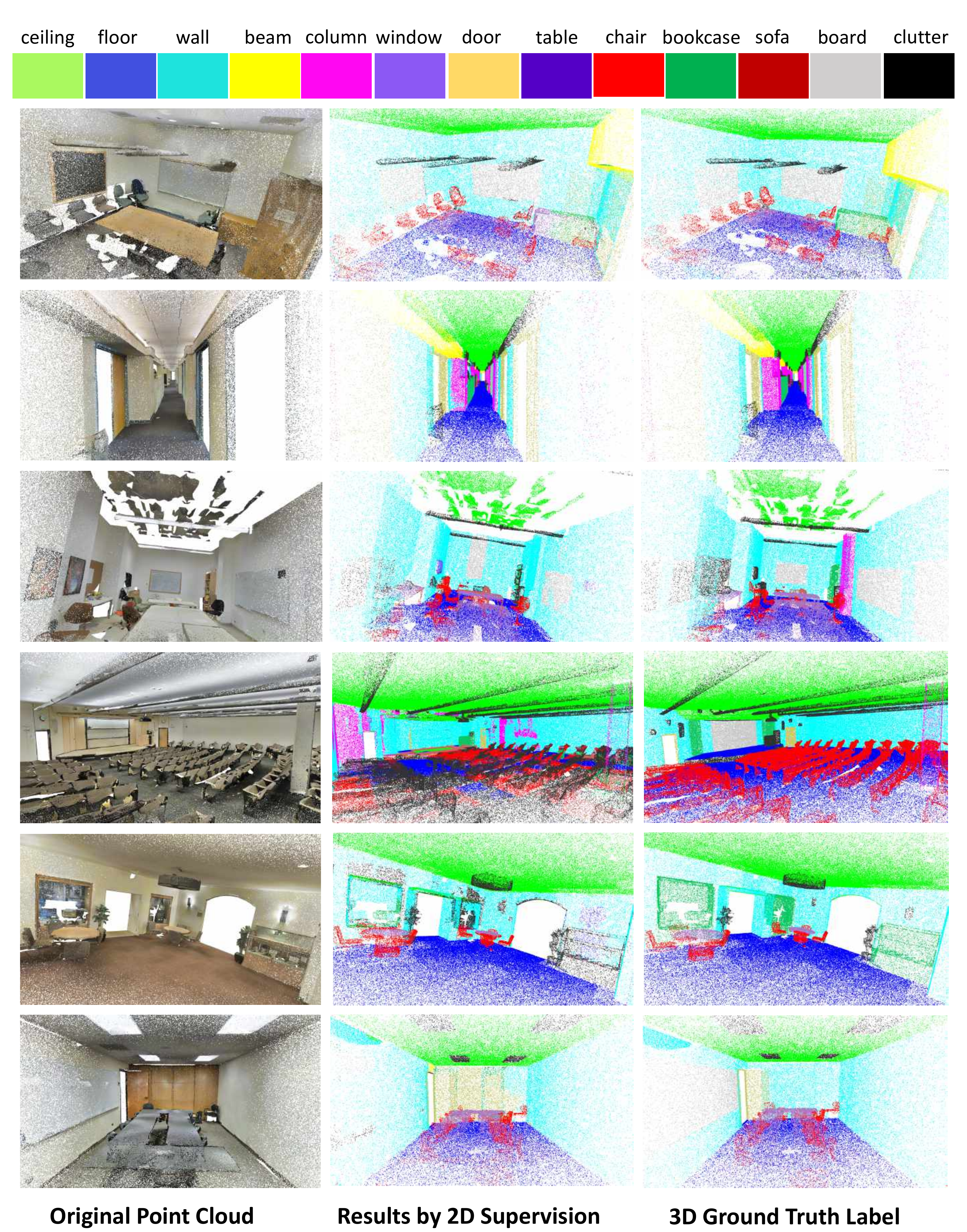}
\caption{Qualitative results produced by our proposed method on S3DIS dataset. The first column is the original point cloud in RGB format. The middle column is the segmentation results by our proposed 2D weakly supervised method. The last column is the ground truth segmentation point cloud for comparison. Overall, our method performs well in most scenes. However, when the scale of the scene is too large and contains crowded objects, the spatial relation and occlusion situation become more complicated which lead to a deficient performance such as the fourth row with many chairs in the scene.}
\label{fig:results}
\end{center}
\end{figure*}

\subsubsection{Comparison with the State-of-the-art Methods}

Since there is no previous work of 2D supervised point cloud semantic segmentation for large-scale natural scenes, here, we compare our proposed framework directly with the state-of-the-arts of fully 3D supervised point cloud segmentation. As shown in Table~\ref{table:comparison_1}, by using only 2D segmentation maps, our method attains comparable results to most of the 3D supervised methods. Note that it even outperforms 3D fully supervised PointNet~\cite{Qi:2017:PointNetDL}. The most recent top-performing 3D point cloud segmentation model, SPG~\cite{Landrieu:2018:LargeScalePC}, still leads a margin in terms of \textit{mean IoU} by applying a hierarchical architecture based on SuperPoints. However, the proposed approach achieves competitive performance in terms of \textit{mean Accuracy} and \textit{overall Accuracy}, without utilizing contextual relationship reasoning as in SPG.

Figures~\ref{fig:results_suncg} and ~\ref{fig:results} visualize several example results on 3D point cloud semantic segmentation generated by our method on SUNCG and S3DIS respectively. Overall, our proposed 2D supervised semantic segmentation method works well in various kinds of areas and rooms contains multiple classes of objects. 

\subsection{Ablation Study}
In this section, we conduct a set of experiments to explore the effects of different encoder designs and various amounts of training data, as well as the accuracy of the visibility detection by OBSNet.

\subsubsection{Encoder Design}
\begin{table}
\begin{center}
\caption{Effects of encoder structures on S3DIS dataset.}
\scalebox{0.81}{
\begin{tabular}{c|c|c|c|c}
\hline
\textbf{K-NN Graph} & \textbf{Pyramid} & \textbf{mAcc(\%)} & \textbf{mIoU(\%)} & \textbf{oAcc(\%)} \\ \hline \hline
\texttimes& \texttimes               &   61.3     &    45.1       &     72.6      \\ 
\checkmark& \texttimes               &   65.1    &     48.6     &      78.4    \\ 
\texttimes& \checkmark            &   63.5     &    46.4       &  75.3         \\ 
\checkmark& \checkmark            &  66.5      &   50.8       &   79.1       \\ 
\end{tabular}
}
\label{table:s3d-network}
\end{center}
\vspace{-1em}
\end{table}

Our GPFN encoder network integrated K-NN graph structure and pyramid design. Here we conduct experiments to verify the effectiveness of these two designs for 3D point cloud semantic segmentation by still using the 2D segmentation maps as the supervision signal. As shown in Table~\ref{table:s3d-network}, without any specific design and extra training data, by using a simple pointnet~\cite{Qi:2017:PointNetDL} similar model, it achieves ($61.3\%$ mAcc, $45.1\%$ mIoU, and $72.6\%$ oAcc) on S3DIS dataset, which show the limited semantic encoding capability with the simple network design. By adding the K-NN graph structure to the encoder network, the performance is boosted up to ($65.1\%$ mAcc, $48.6\%$ mIoU, and $78.4\%$ oAcc) which demonstrate the benefit of using graph convolution to encode the sparse point cloud data. Through the K-NN edge graph convolution, the edge features are extracted and aggregated to the central point. It helps to improve the classify accuracy to the point and to compensate the using 2D supervision signal. By applying the pyramid design which concatenates the global features in both low-level and high-level, the performance increases to ($63.5\%$ mAcc, $46.4\%$ mIoU, and $75.3\%$ oAcc). By adding both the K-NN graph structure and pyramid design, the performance is boosted up to ($66.5\%$ mAcc, $50.8\%$ mIoU, and $79.1\%$ oAcc). This proves that with integrating K-NN graph structure and pyramid design, the network is able to encode more semantic and context information and achieves better segmentation results.

\begin{table}[ht!]
  \centering
  \caption{Performance comparison of using different amount of training data on S3DIS dataset.}
  \setlength{\tabcolsep}{6pt}
  \begin{tabular}{c|c|c|c}
  \hline
    \textbf{Training data} & \textbf{mAcc (\%)} & \textbf{mIoU (\%)} & \textbf{oAcc (\%)}\\ 
    \hline
    \hline
    \textit{All}          &  67.0    &  52.5    &  81.5   \\ 
    \textit{$1/2$}          & 66.9     &  51.8    &  80.9   \\ 
    \textit{$1/4$}          & 66.7     & 50.9     & 79.5    \\ 
    \textit{$1/6$}          & 66.5     & 50.8     & 79.1    \\ 
    \textit{$1/12$}         & 56.5     & 39.3     & 66.2    \\
    \textit{$1/20$}         & 37.8     & 29.1     & 40.0    \\ 
  \end{tabular}
  \\[4pt]
  
  \label{table:proportion}
\end{table}

\begin{figure*}
\centering
\includegraphics[width=0.9\textwidth]{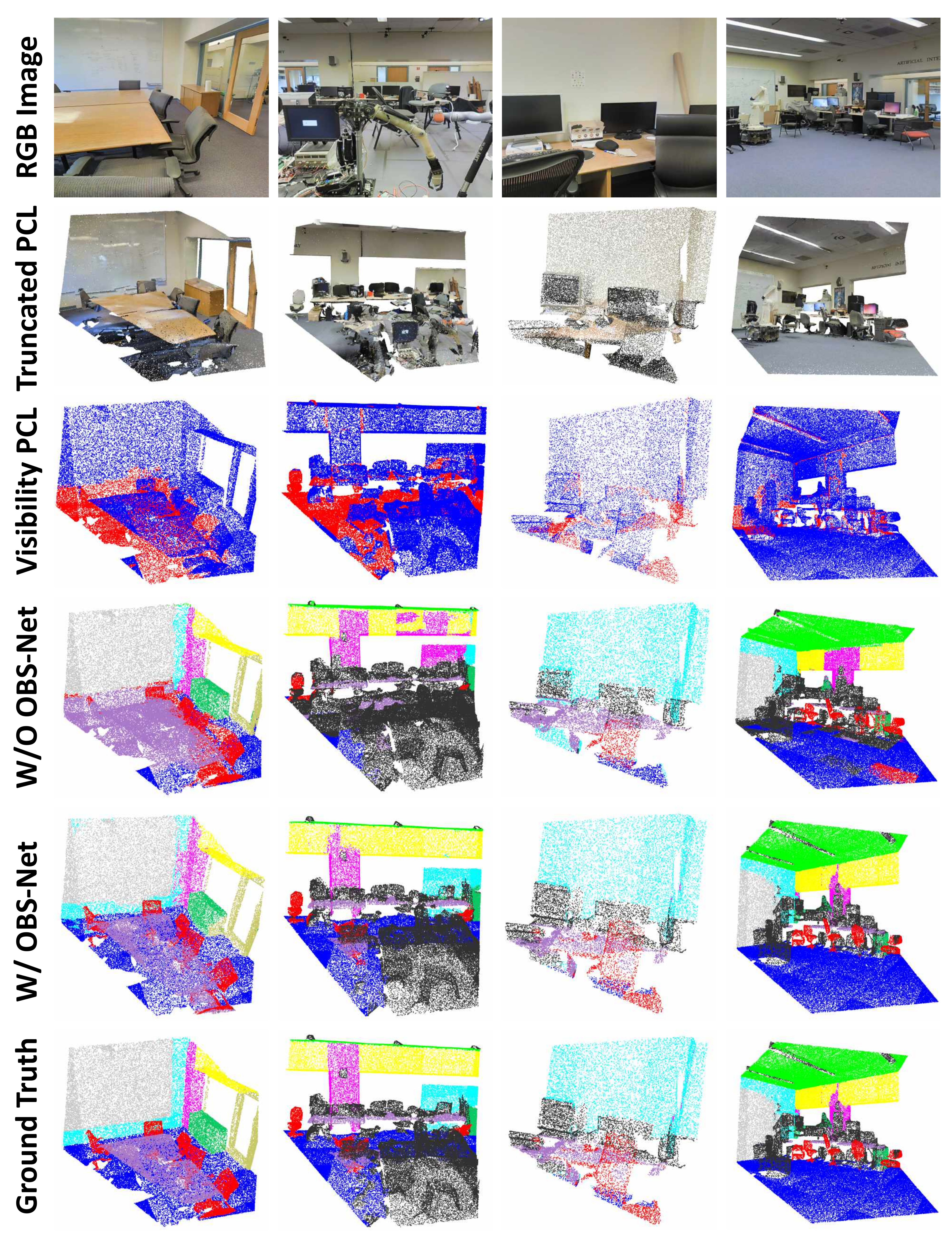}
\caption{Comparison of the segmentation results for several scenes tested on the S3DIS dataset. PCL indicates the point cloud. The first row is the related 2D RGB images (for visualization only, not used in our framework) under a specific viewpoint $v$. The second row shows our truncated point cloud which is fed as the input of our network. The third row demonstrates the output of the OBSNet under the viewpoint $v$. The point cloud is spin for a better visualization. The blue points are the visible part while the red points are the occluded points (only for the third row). The 4th and 5th rows compare the segmentation results with and without the OBSNet. The last row is the ground truth segmentation of 3D point cloud.}
\label{fig:ablation_vis}
\end{figure*}

\subsubsection{Amount of Training Data}
\label{sec:amount}
The scene point clouds of the S3DIS dataset are constructed by thousands of viewpoints. Here the robustness of the proposed point cloud segmentation network is evaluated by using a different amount of training data. Table~\ref{table:proportion} demonstrates the performance of using various data proportion (all, $1/2, 1/4, 1/6, 1/12, 1/20$ of the viewpoints, they are evenly random selected in each room). There is no significant difference between using $1/4$ and $1/6$ of all viewpoints. When using the full scale or $1/2$ training data, due to large-scale training data, the performance is boosted up a little. However, the performance is significantly decreased when using only $1/12$ or $1/20$ viewpoints. That is because when only a few viewpoints are used, some objects might be missed and the occluded objects are hard to be visible in another viewpoint. To balance the trade-off between efficiency and accuracy, $1/6$ data is adopted to conduct all other experiments.

\begin{table}[ht]
\centering
\caption{Accuracy of visibility detection by our proposed OBSNet using different amount of training data on S3DIS dataset.}
\begin{tabular}{c|c|c|c|c|c|c}
\hline
\multirow{2}{*}{\textbf{Dataset}} & \multicolumn{6}{c}{\textbf{Accuracy (\%)}}                                                 \\ \cline{2-7} 
                                  & \textbf{All} & \textbf{1/2} & \textbf{1/4} & \textbf{1/6} & \textbf{1/12} & \textbf{1/20} \\ \hline \hline
\textbf{S3DIS}                    &    93.0       & 92.6         & 91.7         & 91.2         & 89.6          & 85.0         
\end{tabular}

\label{table:vis}
\end{table}

\subsubsection{Visibility Detection by OBSNet}

 As a binary classification, the OBSNet can achieve over 90\% accuracy for visibility detection compared to using distance filter. We train the OBSNet with visibility labels generated by the distance filter and quantitatively evaluate our models on S3DIS datasets, with corresponding results reported in Table~\ref{table:vis}. Given the truncated point cloud as input, the OBSNet first classifies each point as "visible" or "occluded". Following the training data setting in Section~\ref{sec:amount} for a fair comparison, we demonstrate the testing performance of OBSNet by using different amounts of training data. As shown in Table~\ref{table:vis}, there is only $8.0\%$ performance gap between using all and $1/20$ of training data. Even with only $1/20$ of data, the proposed model still achieves $85\%$ of classification accuracy. This further supports our observation that the point clouds between different viewpoints within a room are considerably overlapped with each other. Therefore, reducing training data does not significantly decrease the accuracy. The results show that OBSNet is notably robust to various data amounts of point cloud. 

The effectiveness of the OBSNet is demonstrated in  Figure~\ref{fig:ablation_vis}. As shown in the third row,  the results of the OBSNet, the occluded parts are indicated as red points and the visible parts are visualized as blue points. And through the comparison of the fourth and fifth rows, we observe that the OBSNet successfully separates the visible and occluded objects and improves the segmentation performance. As shown in the first, second, and fourth columns, the occluded parts such as the floor are correctly segmented with the OBSNet. Also in the fourth column, the lights on the ceiling are correctly separated with the visibility detection by the OBSNet. 

\begin{table}[ht]
\centering
\caption{Transfer learning from SUNCG synthetic dataset to S3DIS real-world dataset. First row shows the results on S3DIS trained from scratch without using any pretrained model. Second row shows the results finetuned on S3DIS with the pretrained model on SUNCG dataset.}
\scalebox{0.88}{
\begin{tabular}{c|c|c|c}
\hline
\textbf{Training Data}      & \textbf{mAcc(\%)} & \textbf{mIoU(\%)} & \textbf{oAcc(\%)} \\ \hline \hline
\textbf{Train Scratch on S3DIS}       &   66.5\%            &      50.8\%         &      79.1\%         \\
\textbf{Pretrained on SUNCG } &    67.0\%           &      53.5\%         &      81.3\%        
\end{tabular}
}
\label{table:transfer}
\end{table}

\subsection{Generalization from Synthetic to Real-world}

Since we are the first to explore the semantic point cloud on the SUNCG dataset, there are no other methods to compare. We further explore the domain transfer between the synthetic data to the real-world data to verify the generalization capability of our proposed model. 

First, we pre-train our network on the SUNCG segmentation dataset with the learning rate of $0.005$ and the number of epochs is fixed to 150. The trained features are further finetuned on the S3DIS training dataset. As shown in Table~\ref{table:transfer}, if only train on the S3DIS dataset from scratch, our model achieves ($66.5\%$ mAcc, $50.8\%$ mIoU, and $79.1\%$ oAcc). After adding the pre-trained model on SUNCG, the performance on S3DIS is boosted up to ($67.0\%$ mAcc, $53.5\%$ mIoU, and $81.3\%$ oAcc). Overall the performance consistently improves which demonstrates the generalization capability of our proposed model on real data.

\section{Conclusion}
\label{sec:Conclusion}

In this paper, we have proposed a novel deep graph convolutional model for large-scale semantic scene segmentation in 3D point clouds of wild scenes with only 2D supervision. Combined with the proposed OBSNet and perspective rendering, our proposed method can effectively obtain the semantic segmentation maps of 3D point clouds for both synthetic and real-world scenes. Different from numerous multi-view 2D-supervised methods focusing on only single object point clouds, our proposed method can handle large-scale wild scenes with multiple objects and achieves encouraging performance, with even only a single view per sample. Inferring the occluded part point cloud is the core requirement for the 3D completion task. With the help of semantic information and spatial relation between different objects in the scene, the scene point cloud reconstruction and completion will be benefited from our method. The future directions include unifying the point cloud completion and segmentation tasks for natural scene point clouds.


%


\ifCLASSOPTIONcaptionsoff
  \newpage
\fi



%
\bibliographystyle{plain}      
\bibliography{main}   

\begin{thebibliography}{10}

\bibitem{Achlioptas2017RepresentationLA}
Panos Achlioptas, Olga Diamanti, Ioannis Mitliagkas, and Leonidas~J. Guibas.
\newblock Learning representations and generative models for 3d point clouds.
\newblock In {\em ICML}, 2018.

\bibitem{Armeni20163DSP}
Iro Armeni, Ozan Sener, Amir~Roshan Zamir, Helen Jiang, Ioannis~K. Brilakis,
  Martin~A. Fischer, and Silvio Savarese.
\newblock {3D Semantic Parsing of Large-Scale Indoor Spaces}.
\newblock {\em CVPR}, pages 1534--1543, 2016.

\bibitem{Bithell2007EscapeFT}
Mike Bithell and William~Duncan Macmillan.
\newblock Escape from the cell: Spatially explicit modelling with and without
  grids.
\newblock In {\em International Journal on Ecological Modelling and Systems
  Ecology}, 2007.

\bibitem{Brock2016GenerativeAD}
Andre Brock, Theodore Lim, James~M. Ritchie, and Nick Weston.
\newblock Generative and discriminative voxel modeling with convolutional
  neural networks.
\newblock {\em ArXiv}, abs/1608.04236, 2016.

\bibitem{Chen2019PointBasedMS}
Rui Chen, Songfang Han, Jing Xu, and Hao Su.
\newblock Point-based multi-view stereo network.
\newblock {\em ICCV}, abs/1908.04422, 2019.

\bibitem{Chou2019AVA}
Philip~A. Chou, Maxim Koroteev, and Maja Krivokuca.
\newblock A volumetric approach to point cloud compression—part i: Attribute
  compression.
\newblock {\em IEEE Transactions on Image Processing}, 29:2203--2216, 2019.

\bibitem{Dai:2017:ScanNetR3}
Angela Dai, Angel~Xuan Chang, Manolis Savva, Maciej Halber, Thomas~A.
  Funkhouser, and Matthias Niener.
\newblock {ScanNet: Richly-Annotated 3D Reconstructions of Indoor Scenes}.
\newblock {\em CVPR}, pages 2432--2443, 2017.

\bibitem{Dai2018ScanCompleteLS}
Angela Dai, Daniel Ritchie, Martin Bokeloh, Scott~E. Reed, Jurgen Sturm, and
  Matthias Niener.
\newblock {ScanComplete: Large-Scale Scene Completion and Semantic Segmentation
  for 3D Scans}.
\newblock {\em CVPR}, pages 4578--4587, 2018.

\bibitem{Ding2019PointCS}
Xiaoying Ding, Weisi Lin, Zhenzhong Chen, and Xinfeng Zhang.
\newblock Point cloud saliency detection by local and global feature fusion.
\newblock {\em IEEE Transactions on Image Processing}, 28:5379--5393, 2019.

\bibitem{Engelmann:2017:ExploringSC}
Francis Engelmann, Theodora Kontogianni, Alexander Hermans, and Bastian Leibe.
\newblock {Exploring Spatial Context for 3D Semantic Segmentation of Point
  Clouds}.
\newblock {\em ICCVW}, pages 716--724, 2017.

\bibitem{Engelmann:2018:KnowWY}
Francis Engelmann, Theodora Kontogianni, Jonas Schult, and Bastian Leibe.
\newblock {Know What Your Neighbors Do: 3D Semantic Segmentation of Point
  Clouds}.
\newblock In {\em ECCV Workshops}, 2018.

\bibitem{Garcia2019GeometryCF}
Diogo~C. Garcia, Tiago~A. da~Fonseca, Renan~U. Ferreira, and Ricardo~L.
  de~Queiroz.
\newblock Geometry coding for dynamic voxelized point clouds using octrees and
  multiple contexts.
\newblock {\em IEEE Transactions on Image Processing}, 29:313--322, 2019.

\bibitem{Guerry2017SnapNetRC3}
Joris Guerry, Alexandre Boulch, Bertrand~Le Saux, Julien Moras, Aurelien Plyer,
  and David Filliat.
\newblock {SnapNet-R: Consistent 3D Multi-view Semantic Labeling for Robotics}.
\newblock {\em ICCVW}, pages 669--678, 2017.

\bibitem{He2017MaskR}
Kaiming He, Georgia Gkioxari, Piotr Dollar, and Ross~B. Girshick.
\newblock {Mask R-CNN}.
\newblock {\em ICCV}, pages 2980--2988, 2017.

\bibitem{Hu2019LocalFI}
Wei Hu, Zeqing Fu, and Zongming Guo.
\newblock Local frequency interpretation and non-local self-similarity on graph
  for point cloud inpainting.
\newblock {\em IEEE Transactions on Image Processing}, 28:4087--4100, 2019.

\bibitem{Hua2016SceneNNAS}
Binh-Son Hua, Quang-Hieu Pham, Duc~Thanh Nguyen, Minh-Khoi Tran, Lap-Fai Yu,
  and Sai-Kit Yeung.
\newblock {SceneNN: A Scene Meshes Dataset with aNNotations}.
\newblock {\em 3DV}, pages 92--101, 2016.

\bibitem{Insafutdinov2018UnsupervisedLO}
Eldar Insafutdinov and Alexey Dosovitskiy.
\newblock {Unsupervised Learning of Shape and Pose with Differentiable Point
  Clouds}.
\newblock In {\em NeurIPS}, 2018.

\bibitem{Krivokuca2019AVA}
Maja Krivokuca, Philip~A. Chou, and Maxim Koroteev.
\newblock A volumetric approach to point cloud compression–part ii: Geometry
  compression.
\newblock {\em IEEE Transactions on Image Processing}, 29:2217--2229, 2019.

\bibitem{Landrieu:2018:LargeScalePC}
Loc Landrieu and Martin Simonovsky.
\newblock {Large-Scale Point Cloud Semantic Segmentation with Superpoint
  Graphs}.
\newblock {\em CVPR}, pages 4558--4567, 2018.

\bibitem{Liao20183DSR}
Yi-Lun Liao, Yao-Cheng Yang, and Yu-Chiang~Frank Wang.
\newblock {3D Shape Reconstruction from a Single 2D Image via 2D-3D
  Self-Consistency}.
\newblock {\em CoRR}, abs/1811.12016, 2018.

\bibitem{Lin:2018:LearningEP}
Chen-Hsuan Lin, Chen Kong, and Simon Lucey.
\newblock {Learning Efficient Point Cloud Generation for Dense 3D Object
  Reconstruction}.
\newblock In {\em AAAI}, 2018.

\bibitem{Mandikal:2018:3DLMNetLE}
Priyanka Mandikal, L.~NavaneetK., Mayank Agarwal, and Venkatesh~Babu
  Radhakrishnan.
\newblock {3D-LMNet: Latent Embedding Matching for Accurate and Diverse 3D
  Point Cloud Reconstruction from a Single Image}.
\newblock In {\em BMVC}, 2018.

\bibitem{Maturana2015VoxNetA3}
Daniel Maturana and Sebastian~A. Scherer.
\newblock Voxnet: A 3d convolutional neural network for real-time object
  recognition.
\newblock {\em 2015 IEEE/RSJ International Conference on Intelligent Robots and
  Systems (IROS)}, pages 922--928, 2015.

\bibitem{Meng2018VVNetVV}
Hsien-Yu Meng, Lin Gao, Yu-Kun Lai, and Dinesh Manocha.
\newblock Vv-net: Voxel vae net with group convolutions for point cloud
  segmentation.
\newblock {\em 2019 IEEE/CVF International Conference on Computer Vision
  (ICCV)}, pages 8499--8507, 2018.

\bibitem{NavaneetK:2019:CAPNetCA}
L~NavaneetK, Priyanka Mandikal, Mayank Agarwal, and R.~Venkatesh Babu.
\newblock {CAPNet: Continuous Approximation Projection For 3D Point Cloud
  Reconstruction Using 2D Supervision}.
\newblock {\em CoRR}, abs/1811.11731, 2019.

\bibitem{Pittaluga2019RevealingSB}
Francesco Pittaluga, Sanjeev~J. Koppal, Sing~Bing Kang, and Sudipta~N. Sinha.
\newblock Revealing scenes by inverting structure from motion reconstructions.
\newblock In {\em CVPR}, 2019.

\bibitem{qi2017frustum}
Charles~R. Qi, Wei Liu, Chenxia Wu, Hao Su, and Leonidas~J. Guibas.
\newblock Frustum pointnets for 3d object detection from rgb-d data.
\newblock {\em 2018 IEEE/CVF Conference on Computer Vision and Pattern
  Recognition}, pages 918--927, 2017.

\bibitem{Qi:2017:PointNetDL}
Charles~R Qi, Hao Su, Kaichun Mo, and Leonidas~J. Guibas.
\newblock {PointNet: Deep Learning on Point Sets for 3D Classification and
  Segmentation}.
\newblock In {\em CVPR}, 2017.

\bibitem{Qi:2017:PointNetDH}
Charles~R Qi, Li~Yi, Hao Su, and Leonidas~J. Guibas.
\newblock {PointNet++: Deep Hierarchical Feature Learning on Point Sets in a
  Metric Space}.
\newblock In {\em NIPS}, 2017.

\bibitem{Qi2016VolumetricAM}
Charles~Ruizhongtai Qi, Hao Su, Matthias Niener, Angela Dai, Mengyuan Yan, and
  Leonidas~J. Guibas.
\newblock {Volumetric and Multi-view CNNs for Object Classification on 3D
  Data}.
\newblock {\em CVPR}, pages 5648--5656, 2016.

\bibitem{Riegler2016OctNetLD}
Gernot Riegler, Ali~O. Ulusoy, and Andreas Geiger.
\newblock Octnet: Learning deep 3d representations at high resolutions.
\newblock {\em 2017 IEEE Conference on Computer Vision and Pattern Recognition
  (CVPR)}, pages 6620--6629, 2016.

\bibitem{Shi2015DeepPanoDP}
Baoguang Shi, Song Bai, Zhichao Zhou, and Xiang Bai.
\newblock {DeepPano: Deep Panoramic Representation for 3-D Shape Recognition}.
\newblock {\em IEEE Signal Processing Letters}, 22:2339--2343, 2015.

\bibitem{Song2017SemanticSC}
Shuran Song, Fisher Yu, Andy Zeng, Angel~Xuan Chang, Manolis Savva, and
  Thomas~A. Funkhouser.
\newblock {Semantic Scene Completion from a Single Depth Image}.
\newblock {\em CVPR}, pages 190--198, 2017.

\bibitem{Su2015MultiviewCN}
Hang Su, Subhransu Maji, Evangelos Kalogerakis, and Erik~G. Learned-Miller.
\newblock {Multi-view Convolutional Neural Networks for 3D Shape Recognition}.
\newblock {\em ICCV}, pages 945--953, 2015.

\bibitem{Tchapmi2017SEGCloudSS}
Lyne~P. Tchapmi, Christopher~Bongsoo Choy, Iro Armeni, JunYoung Gwak, and
  Silvio Savarese.
\newblock Segcloud: Semantic segmentation of 3d point clouds.
\newblock {\em 2017 International Conference on 3D Vision (3DV)}, pages
  537--547, 2017.

\bibitem{Te:2018:RGCNNRG}
Gusi Te, Wei Hu, Amin Zheng, and Zongming Guo.
\newblock {RGCNN: Regularized Graph CNN for Point Cloud Segmentation}.
\newblock In {\em ACM Multimedia}, 2018.

\bibitem{Wang2019LDLS3O}
Brian~H. Wang, Wei-Lun Chao, Yulin Wang, Bharath Hariharan, Kilian~Q.
  Weinberger, and Mark~E. Campbell.
\newblock Ldls: 3-d object segmentation through label diffusion from 2-d
  images.
\newblock {\em IEEE Robotics and Automation Letters}, 4:2902--2909, 2019.

\bibitem{HaiyanBMVC}
Haiyan Wang, Xuejian Rong, Liang Yang, Shuihua Wang, and Yingli Tian.
\newblock {Towards Weakly Supervised Semantic Segmentation in 3D
  Graph-Structured Point Clouds of Wild Scenes}.
\newblock In {\em BMVC}, 2019.

\bibitem{Wang:2019:AssociativelySI}
Xinlong Wang, Shu Liu, Xiaoyong Shen, Chunhua Shen, and Jiaya Jia.
\newblock {Associatively Segmenting Instances and Semantics in Point Clouds}.
\newblock {\em CoRR}, abs/1902.09852, 2019.

\bibitem{Wang:2018:DynamicGC}
Yue Wang, Yongbin Sun, Ziwei Liu, Sanjay~E. Sarma, Michael~M. Bronstein, and
  Justin~M. Solomon.
\newblock {Dynamic Graph CNN for Learning on Point Clouds}.
\newblock {\em CoRR}, abs/1801.07829, 2018.

\bibitem{Yang2017FoldingNetPC}
Yaoqing Yang, Chen Feng, Yiru Shen, and Dong Tian.
\newblock Foldingnet: Point cloud auto-encoder via deep grid deformation.
\newblock {\em 2018 IEEE/CVF Conference on Computer Vision and Pattern
  Recognition}, pages 206--215, 2017.

\bibitem{Ye20183DRN}
Xiaoqing Ye, Jiamao Li, Hexiao Huang, Liang Du, and Xiaolin Zhang.
\newblock {3D Recurrent Neural Networks with Context Fusion for Point Cloud
  Semantic Segmentation}.
\newblock In {\em ECCV}, 2018.

\bibitem{Yuan2018PCNPC}
Wentao Yuan, Tejas Khot, David Held, Christoph Mertz, and Martial Hebert.
\newblock {PCN: Point Completion Network}.
\newblock In {\em 3DV}, 2018.

\end{thebibliography}

%




\end{document}